\documentclass{article}

\usepackage[preprint]{neurips_2026}

\usepackage[utf8]{inputenc}
\usepackage[T1]{fontenc}
\usepackage{hyperref}
\usepackage{url}
\usepackage{booktabs}
\usepackage{amsfonts}
\usepackage{amsmath}
\usepackage{amssymb}
\usepackage{nicefrac}
\usepackage{microtype}
\usepackage{graphicx}
\graphicspath{{figures/}}
\graphicspath{{figures/}}
\usepackage{xcolor}
\usepackage{subcaption}
\usepackage{multirow}
\usepackage{enumitem}
\usepackage{algorithm}
\usepackage{algorithmic}

\newcommand{\ours}{\textsc{oxRL}}

\title{Do Post-Training Algorithms Actually Differ? \\A Controlled Study Across Model Scales Uncovers Scale-Dependent Ranking Inversions}

\author{
  Xiaoyi Li \\
  \texttt{orze.ai@hotmail.com}
}

\begin{document}

\maketitle

\begin{abstract}
Post-training alignment has produced dozens of competing algorithms---DPO, SimPO, KTO, GRPO, and others---yet practitioners lack controlled comparisons to guide algorithm selection.
We present \ours{}, a unified framework implementing 51 post-training algorithms with identical infrastructure, enabling the first large-scale apples-to-apples evaluation.
Our study spans \textbf{8 algorithms} across \textbf{4 model scales} (0.5B--7B), \textbf{3 evaluation domains}, and a \textbf{20-variant DPO taxonomy} (100 runs at 1.5B, 5 seeds each), totaling $\sim$240 training runs on H100 GPUs.
Three headline findings emerge.
\textbf{(1)~Algorithm rankings are unstable across scale}: at 1.5B, online RL (SGRPO) tops all methods at 58.0\%~$\pm$0.57 on GSM8K; by 7B, the worst small-scale method (SimPO) becomes the best (85.8\%), a complete ranking inversion driven by model scale rather than LoRA regularization (confirmed via 2$\times$2 factorial).
\textbf{(2)~Loss function modifications yield negligible gains}: none of 20 DPO variants significantly outperform vanilla DPO after Bonferroni correction; the sole significant outlier, SimPO, is \emph{worse} ($-$11.5~pp, $p < 10^{-4}$).
\textbf{(3)~Algorithm leverage is task-specific}: the 19.3~pp GSM8K spread collapses to 0.54~pp on MATH ($36\times$) and 0.47~pp on general-domain benchmarks ($41\times$), confirming that algorithm choice matters primarily within the training distribution.
These findings yield a hierarchy of leverage for practitioners: model scale (${\sim}$50~pp) $\gg$ training paradigm (${\sim}$10~pp) $\gg$ online vs.\ offline (${\sim}$9~pp) $\gg$ loss function (${\sim}$1~pp).
We release all code, configs, and evaluation data as a living community benchmark.
\end{abstract}

\section{Introduction}

Post-training of large language models (LLMs)---aligning a pretrained model with human preferences or task objectives---has become the decisive stage separating a capable base model from a useful one~\citep{ouyang2022training, touvron2023llama2}.
The past two years have produced an explosion of post-training algorithms: online RL methods like PPO~\citep{schulman2017proximal} and GRPO~\citep{shao2024deepseekmath}, offline preference methods like DPO~\citep{rafailov2023direct}, and a growing family of variants---SimPO~\citep{meng2024simpo}, KTO~\citep{ethayarajh2024kto}, ORPO~\citep{hong2024orpo}---each claiming improvements.
DeepSeek-R1~\citep{deepseek2025r1} demonstrated that pure RL post-training can elicit sophisticated reasoning, further expanding the design space.

\textbf{The practitioner's dilemma.}
Each algorithm paper reports results on different base models, datasets, and evaluation suites, making cross-method comparison unreliable.
The few existing comparative studies~\citep{xu2024dpo} cover only 2--3 methods and often confound algorithmic differences with implementation details.
Three critical dimensions remain unexplored: how rankings change across \emph{model scale}, which DPO variant modifications actually matter (over 20 exist~\citep{rafailov2023direct}), and the compute-performance tradeoff between online and offline methods.

\textbf{Our approach.}
We address these gaps through the first large-scale, controlled empirical comparison of post-training algorithms, resting on three pillars:
\begin{enumerate}[leftmargin=*,itemsep=1pt]
    \item \textbf{Unified framework.} All 51 algorithms share the same model loading, data pipeline, distributed training (DeepSpeed ZeRO-3), and evaluation harness, eliminating codebase-as-confound.
    \item \textbf{Controlled design.} We fix the model family (Qwen 2.5), training data, optimizer, and LR schedule, varying only the loss function. Where algorithms require additional components (reference models, rollouts), these are standardized.
    \item \textbf{Multi-scale evaluation.} 8 algorithms across 4 scales (0.5B, 1.5B, 3B, 7B) on GSM8K, plus 20 DPO variants at 1.5B with 5 seeds each. At 3B, we train under both full FT and LoRA, providing a 2$\times$2 factorial with 7B to disentangle scale from LoRA effects.
\end{enumerate}

\textbf{Key findings.}
\begin{itemize}[leftmargin=*,itemsep=1pt]
    \item \textbf{Rankings invert across scale.} At 1.5B, SGRPO (online RL) tops all methods at 58.0\%~$\pm$0.57, outperforming SFT by 3.6~pp and DPO by 8.9~pp. By 7B, SimPO---worst at small scale---becomes best (85.8\%), while SFT collapses to near-baseline. A 2$\times$2 factorial confirms scale, not LoRA, drives the inversion.
    \item \textbf{Zero of 20 DPO variants beat vanilla DPO.} In 100 runs with Bonferroni correction, SimPO is the sole significant outlier ($-$11.5~pp, $p < 10^{-4}$)---and it is \emph{worse}.
    \item \textbf{Algorithm leverage is task-specific.} The 19.3~pp GSM8K spread collapses to 0.54~pp on MATH ($36\times$) and 0.47~pp on general-domain benchmarks ($41\times$); no method degrades general capabilities.
    \item \textbf{A hierarchy of leverage.} Scale (${\sim}$50~pp) $\gg$ training paradigm (${\sim}$10~pp) $\gg$ online vs.\ offline (${\sim}$9~pp) $\gg$ loss function (${\sim}$1~pp).
    \item \textbf{A methodological warning.} We discover a hidden determinism bug in PyTorch's \texttt{Distributed\-Sampler} that silently eliminates seed-dependent variance, likely affecting other distributed training studies.
\end{itemize}

We release \ours{} as a standardized benchmark protocol for post-training algorithms---analogous to GLUE for NLU or HELM for evaluation, but targeting the \emph{training algorithm} axis.\footnote{Code and data: \url{https://github.com/warlockee/oxRL}.}
We provide four artifacts: (1)~a codebase where any new algorithm can be compared by implementing a single loss function; (2)~a prescribed evaluation protocol with statistical tests; (3)~reproducible YAML configs for every algorithm~$\times$~scale combination; and (4)~all raw evaluation outputs for independent verification.

\section{Related Work}

\textbf{RLHF and online RL.}
The RLHF pipeline~\citep{ouyang2022training, touvron2023llama2} trains a reward model and optimizes the policy using PPO~\citep{schulman2017proximal} with a KL penalty.
GRPO~\citep{shao2024deepseekmath} simplifies this by eliminating the value network through group-normalized rewards.
We evaluate three GRPO loss variants (SGRPO, GSPO, CISPO), showing that the token-vs-sentence-level loss distinction exhibits scale dependence.

\textbf{DPO and its variants.}
\citet{rafailov2023direct} showed the RLHF objective can be reparameterized to bypass reward modeling.
This spawned a large family: IPO~\citep{azar2024general}, SimPO~\citep{meng2024simpo}, KTO~\citep{ethayarajh2024kto}, ORPO~\citep{hong2024orpo}, and others.
Our work provides the first controlled evaluation of 20 DPO variants under identical conditions.

\textbf{Comparative studies.}
\citet{xu2024dpo} compare DPO and PPO, and \citet{ivison2024unpacking} disentangle data, procedures, and loss functions, but both cover only 2--3 algorithms.
\citet{spangher2025rlhf} benchmark 17 algorithms across 3,500 runs but at a single scale---they cannot detect the ranking inversions we observe between 3B and 7B.
\citet{wu2025grpo_dpo} demonstrate GRPO--DPO theoretical equivalence; our empirical results partially contradict this (SGRPO outperforms DPO by 8.9~pp at 1.5B).
\citet{saeidi2025dpo_variants} evaluate DPO variants without multi-seed runs or multiple comparison correction; our 100-run Bonferroni-corrected sweep provides stronger statistical evidence.
\citet{tajwar2024preference} and \citet{tan2025rl_scaling} address related dimensions (on-policy data, RL scaling laws) that our work complements.
No prior work explores the DPO variant space at our scale or examines ranking stability across model sizes.
Our 0/20 result for DPO variants parallels the finding of \citet{lucic2018gans} that most GAN variants do not outperform the original under controlled evaluation---a pattern that may be general to families of loss function modifications.

\textbf{Frameworks and benchmarks.}
TRL~\citep{vonwerra2022trl}, OpenRLHF, and DeepSpeed-Chat focus on training infrastructure; HELM~\citep{liang2023holistic} and Open LLM Leaderboard provide evaluation but do not control training conditions.
The Zephyr~\citep{tunstall2023zephyr} and Alignment Handbook~\citep{lambert2024alignment} projects provide practical recipes for single pipelines.
Scaling law studies~\citep{kaplan2020scaling, hoffmann2022training} characterize pretraining but not post-training.
\ours{} is designed as an \emph{evaluation benchmark} that enforces identical conditions across methods with statistical methodology for rigorous comparison.

\section{Methods}
\label{sec:methods}

\subsection{Framework Design}
\label{sec:framework}

\ours{} provides a single framework where all 51 algorithms share identical model loading (HuggingFace, \texttt{flash\_attention\_2}), data pipelines (prompt-only, prompt-response, prompt-preference), distributed training (DeepSpeed ZeRO-3~\citep{rajbhandari2020zero}, vLLM~\citep{kwon2023efficient} for RL rollouts), and checkpointing.
Each algorithm implements a \texttt{BaseAlgorithm} interface requiring only the loss function.
This ensures the \emph{only} difference between two SL runs is the loss computation---model, data order, optimizer state, and gradient pipeline are byte-identical up to the loss backward pass.

\subsection{Algorithm Taxonomy}
\label{sec:taxonomy}

We organize the 51 algorithms into four families:
\textbf{(1) SFT:} next-token prediction on curated responses.
\textbf{(2) Online RL:} PPO~\citep{schulman2017proximal} and GRPO~\citep{shao2024deepseekmath} with three loss variants (SGRPO, GSPO, CISPO).
\textbf{(3) Offline preference optimization:} methods learning from pre-collected pairs $(x, y_w, y_l)$---DPO~\citep{rafailov2023direct}, IPO~\citep{azar2024general}, SimPO~\citep{meng2024simpo}, KTO~\citep{ethayarajh2024kto}, and 16 additional variants (Table~\ref{tab:dpo_variants}).
\textbf{(4) Hybrid:} ORPO~\citep{hong2024orpo}, Online DPO, SPIN~\citep{chen2024spin}.
The core loss functions are:
\begin{align}
    \text{DPO:}& \quad \mathcal{L} = -\log\sigma\!\left(\beta\left[\log\frac{\pi_\theta(y_w|x)}{\pi_{\text{ref}}(y_w|x)} - \log\frac{\pi_\theta(y_l|x)}{\pi_{\text{ref}}(y_l|x)}\right]\right) \label{eq:dpo} \\
    \text{IPO:}& \quad \mathcal{L} = \left(\log\frac{\pi_\theta(y_w|x)}{\pi_{\text{ref}}(y_w|x)} - \log\frac{\pi_\theta(y_l|x)}{\pi_{\text{ref}}(y_l|x)} - \frac{1}{2\beta}\right)^2 \label{eq:ipo} \\
    \text{SimPO:}& \quad \mathcal{L} = -\log\sigma\!\left(\frac{\beta}{|y_w|}\log\pi_\theta(y_w|x) - \frac{\beta}{|y_l|}\log\pi_\theta(y_l|x) - \gamma\right) \label{eq:simpo}
\end{align}
where $\sigma$ is the sigmoid, $\beta$ controls preference signal strength, and $\gamma$ is SimPO's target margin.
Our 20-variant DPO taxonomy (Table~\ref{tab:dpo_variants}) categorizes modifications along axes including divergence measure, reference model usage, length normalization, weighting schemes, and regularization strategies.

\begin{table}[t]
\centering
\caption{Taxonomy of 20 DPO variants evaluated in this study, organized by primary modification to the DPO objective (Eq.~\ref{eq:dpo}). All evaluated at 1.5B on GSM8K with 5 seeds.}
\label{tab:dpo_variants}
\small
\begin{tabular}{@{}llcc@{}}
\toprule
\textbf{Variant} & \textbf{Primary Modification} & \textbf{Ref.} & \textbf{Year} \\
\midrule
DPO & Baseline logistic loss & \checkmark & 2023 \\
IPO & Squared loss regularization & \checkmark & 2024 \\
SimPO & Length-normalized, no ref & $\times$ & 2024 \\
KTO & Unpaired binary feedback & \checkmark & 2024 \\
Hinge & Hinge loss replacement & \checkmark & 2024 \\
CPO & Contrastive preference & $\times$ & 2024 \\
ORPO & SFT-integrated preference & $\times$ & 2024 \\
RDPO & Robust divergence penalty & \checkmark & 2024 \\
CDPO & Conservative DPO & \checkmark & 2024 \\
BetaDPO & Adaptive $\beta$ scheduling & \checkmark & 2024 \\
CalDPO & Calibrated reward margin & \checkmark & 2024 \\
DPOP & Pref.-weighted optimization & \checkmark & 2024 \\
ODPO & Online DPO & \checkmark & 2024 \\
EXO & Exponentiated gradient & \checkmark & 2024 \\
AlphaPO & Alpha-divergence pref. & $\times$ & 2024 \\
APO & Anchored preference & \checkmark & 2024 \\
SPPO & Self-play preference & \checkmark & 2024 \\
RobustDPO & Dist.\ robust DPO & \checkmark & 2024 \\
GPO & Generalized pref.\ opt. & \checkmark & 2024 \\
FocalPO & Focal loss preference & \checkmark & 2024 \\
\bottomrule
\end{tabular}
\end{table}

\subsection{Evaluation Protocol}
\label{sec:eval_protocol}

Our primary benchmark is GSM8K~\citep{cobbe2021gsm8k} (1,319 test problems, exact-match accuracy with greedy decoding), chosen for its deterministic, verifiable answers.
We additionally evaluate on MATH~\citep{hendrycks2021math} (5,000 problems, 4-shot, minerva-math exact-match) at 1.5B.
All evaluations use the Language Model Evaluation Harness~\citep{eval-harness}: zero-shot chain-of-thought at 0.5B, 8-shot standard at 1.5B and 7B.
For the 20-variant DPO taxonomy, we perform pairwise Welch's $t$-tests against vanilla DPO with Bonferroni correction ($\alpha = 0.05/19 \approx 0.0026$).
We release configurations for MBPP~\citep{austin2021program} for community extension.

We discovered post-hoc that our SL training pipeline is fully deterministic across seeds due to a \texttt{DistributedSampler} seed propagation issue (Appendix~\ref{app:seed_invariance}).
After fixing seed propagation, we ran verification experiments at both 1.5B (5 algorithms $\times$ 3 seeds = 15 runs) and 3B (5 algorithms, 1--2 seeds each), confirming all rankings while revealing genuine variance ($\sigma$ up to 2.01~pp at 3B).
We report seed-fix means with standard deviations where multi-seed data is available, and mark single-seed values with $^\ddagger$ (Appendix~\ref{app:seed_invariance}).

\section{Experimental Setup}

Our experiments comprise three studies: (1) \textbf{core comparison} of 8 algorithms across 4 scales ($\sim$55 runs), including a 3B full-FT/LoRA factorial; (2) \textbf{DPO variant taxonomy} (20 variants $\times$ 5 seeds = 100 runs at 1.5B); and (3) \textbf{ablations} ($\sim$48 runs). Total: $\sim$240 runs requiring $\sim$235 GPU-hours on H100s (Appendix~\ref{app:compute}).

\textbf{Models.} Qwen 2.5 Instruct~\citep{qwen2025qwen25} at 0.5B, 1.5B, 3B, 7B, initialized from Instruct checkpoints~\citep{qwen2025qwen25}.
\textbf{Data.} Preference pairs from model self-play on GSM8K's 7,473 training problems; SFT uses gold responses.
\textbf{Training.} AdamW ($\beta_1\!=\!0.9$, $\beta_2\!=\!0.95$), cosine LR with 10\% warmup, LR $10^{-6}$, 3 epochs over 400 micro-batches per epoch, batch size 8.
LoRA~\citep{hu2022lora} (rank 16, $\alpha\!=\!32$) at 7B and for the 3B factorial; full FT at $\leq$3B for the core comparison.
BF16 on H100 80GB: 1 GPU per SL job, 2 per RL job.
$\beta\!=\!0.1$, SimPO $\gamma\!=\!0.5$, GRPO $n\!=\!4$, $\lambda_{\text{KL}}\!=\!0$.
Full configurations are in Appendix~\ref{app:configs}.

\section{Experiments and Results}

\subsection{Core Algorithm Comparison Across Scales}
\label{sec:core_comparison}

Table~\ref{tab:main_results} presents our main results.
At \textbf{0.5B}, all methods improve over the 19.56\% base model except GSPO (19.86\%) and CISPO (0.15\%, pathological).
IPO leads at 34.50\%, with DPO and SFT tied at 33.97\% and SGRPO close at 32.45\%.
SimPO achieves only 26.08\%, substantially below reference-based methods.

At \textbf{1.5B}, the picture shifts.
SGRPO achieves the highest accuracy: $58.00\% \pm 0.57$ (3 seeds), surpassing SFT ($54.36\% \pm 0.59$) by 3.6~pp and DPO ($49.08\% \pm 0.61$) by 8.9~pp---the largest single-method improvement at this scale.
Among offline methods, the ranking holds: SFT (54.4\%) $>$ IPO (52.2\%) $>$ KTO (51.2\%) $>$ DPO (49.1\%) $>$ SimPO (38.7\%), with SFT outperforming all preference methods.
The reference-based methods (DPO, IPO, KTO) cluster within 3.2~pp, consistent with the preference loss function being a low-impact choice.
SimPO's deficit widens to $-$15.7~pp vs.\ SFT, foreshadowing the scale-dependent reversal at 7B.

At \textbf{3B}, accuracy \emph{drops} for all methods compared to 1.5B due to severe format non-compliance: the base model achieves 62.9\% with flexible extraction but only 10.69\% strict-match.
SFT leads at 18.35\%, followed by KTO (15.31\%), DPO (14.44\% $\pm$ 2.01), and SimPO (6.90\%---below baseline).
DPO exhibits the highest seed variance ($\sigma = 2.01$~pp vs.\ $\sigma = 0.61$ at 1.5B), indicating scale-dependent training instability.
The ranking at 3B mirrors 1.5B: SFT $>$ reference-based methods $>$ SimPO.

\begin{figure}[t]
\centering
\includegraphics[width=0.8\textwidth]{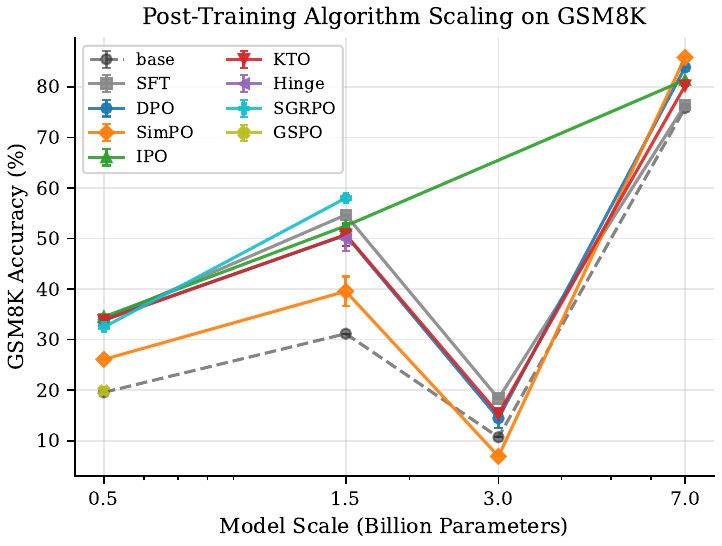}
\caption{GSM8K accuracy across model scales. Dashed gray: base model. Error bars: $\pm$1$\sigma$ where multi-seed data is available. The V-shaped trajectories from 3B to 7B visualize the scale-driven ranking inversion; DPO's error bar at 3B reflects genuine seed variance ($\sigma = 2.01$~pp). At 1.5B, SGRPO achieves the highest accuracy (58.0\%). Note: 0.5B--3B use full FT; 7B uses LoRA.}
\label{fig:scaling_curves}
\end{figure}

At \textbf{7B}, the ordering \emph{completely reverses} (Figure~\ref{fig:scaling_curves}): SimPO (85.82\%) $>$ DPO (83.85\%) $>$ IPO (81.35\%) $>$ KTO (80.21\%) $>$ SFT (76.42\%).
SimPO swings from worst to best---a $\sim$21~pp reversal relative to SFT.
SFT collapses to near-baseline (+0.6~pp over base).

\textbf{Disentangling scale from LoRA.}
Since all 7B experiments use LoRA, the inversion could reflect LoRA's regularization rather than scale.
Table~\ref{tab:lora_2x2} presents our 2$\times$2 factorial (3B/7B $\times$ full FT/LoRA).
At 3B, LoRA has \emph{no detectable effect}: SFT differs by only 0.31~pp (18.35\% full FT vs.\ 18.04\% LoRA), and the DPO gap (1.56~pp) is well within DPO's seed variance ($\sigma = 2.01$).
SimPO \emph{degrades below baseline} at 3B under both methods ($-$3.8~pp).
Since LoRA is held constant between the 3B and 7B rows, the massive accuracy jumps---DPO: 16.0\%$\to$83.9\% (+67.9~pp), SFT: 18.0\%$\to$76.4\% (+58.4~pp)---are attributable to \textbf{model scale}.
The ranking inversion requires sufficient model capacity ($\geq$7B); LoRA alone is insufficient.
Scale does the heavy lifting; LoRA's role is enabling (making 7B training feasible) rather than causal.
Without 7B full fine-tuning (infeasible on our hardware), we cannot fully isolate whether the inversion also occurs under full FT; however, the null LoRA effect at 3B makes it unlikely that LoRA explains the inversion at 7B.
Format compliance contributes to the inversion: SimPO produces exceptionally well-formatted answers at 7B (strict $>$ flexible by 4.9~pp), while SFT conceals 3.9~pp of accuracy behind format-noncompliant outputs (Appendix~\ref{app:format}).

\begin{table}[t]
\centering
\caption{The 2$\times$2 factorial disentangling scale from LoRA. GSM8K strict-match accuracy (\%). Full FT at 3B uses seed-fix reruns: DPO reports mean $\pm \sigma$ over 2 seeds; $^\ddagger$single seed available. At 3B, the full FT/LoRA gap is negligible for all methods (SFT: 0.31~pp, DPO: 1.56~pp---within DPO's seed variance of $\sigma = 2.01$, SimPO: 0.00~pp). The 3B$\to$7B gains (DPO: +67.9~pp, SimPO: +78.9~pp) are attributable to \textbf{scale}. ``---'' = not run (7B full FT).}
\label{tab:lora_2x2}
\small
\begin{tabular}{@{}llcc@{}}
\toprule
& & \textbf{3B} & \textbf{7B} \\
\midrule
\multirow{4}{*}{\textbf{Full FT}} & Base & 10.69 & 75.82 \\
& SFT & \textbf{18.35}$^\ddagger$ \small{(\textcolor{teal}{+7.7})} & --- \\
& DPO & 14.44$_{\pm 2.01}$ \small{(\textcolor{teal}{+3.8})} & --- \\
& SimPO & 6.90$^\ddagger$ \small{(\textcolor{red}{$-$3.8})} & --- \\
\midrule
\multirow{4}{*}{\textbf{LoRA}} & Base & 10.69 & 75.82 \\
& SFT & 18.04 \small{(\textcolor{teal}{+7.4})} & 76.42 \small{(\textcolor{teal}{+0.6})} \\
& DPO & 16.00 \small{(\textcolor{teal}{+5.3})} & 83.85 \small{(\textcolor{teal}{+8.0})} \\
& SimPO & 6.90 \small{(\textcolor{red}{$-$3.8})} & \textbf{85.82} \small{(\textcolor{teal}{+10.0})} \\
\bottomrule
\end{tabular}
\end{table}

\begin{table}[t]
\centering
\caption{GSM8K exact-match accuracy (\%) across scales. 1.5B values are seed-fix means with $\pm \sigma$ across 3 seeds (Appendix~\ref{app:seed_invariance}). 3B uses full FT with seed-fixed reruns; DPO reports mean $\pm \sigma$ over 2 seeds, $^\ddagger$single seed available. LoRA results at 3B appear in Table~\ref{tab:lora_2x2}. Best per scale in \textbf{bold}. $^\dagger$Epoch-1 checkpoint. $^*$Pathological failure. $^\S$62.9\% with flexible extraction but 10.69\% strict-match, indicating format non-compliance. CISPO = Clipped Importance-Sampled Policy Optimization, a sentence-level GRPO variant with per-token ratio clipping.}
\label{tab:main_results}
\small
\begin{tabular}{@{}lcccc@{}}
\toprule
\textbf{Algorithm} & \textbf{0.5B} & \textbf{1.5B} & \textbf{3B (full FT)} & \textbf{7B (LoRA)} \\
\midrule
Base Model & 19.56 & 31.16 & 10.69$^\S$ & 75.82 \\
\midrule
SFT & 33.97$_{\pm 0.0}$ \small{(\textcolor{teal}{+14.4})} & 54.36$_{\pm 0.59}$ \small{(\textcolor{teal}{+23.2})} & \textbf{18.35}$^\ddagger$ \small{(\textcolor{teal}{+7.7})} & 76.42$_{\pm 0.0}$ \small{(\textcolor{teal}{+0.6})} \\
\midrule
DPO & 33.97$_{\pm 0.0}$ \small{(\textcolor{teal}{+14.4})} & 49.08$_{\pm 0.61}$ \small{(\textcolor{teal}{+17.9})} & 14.44$_{\pm 2.01}$ \small{(\textcolor{teal}{+3.8})} & 83.85$_{\pm 0.0}$ \small{(\textcolor{teal}{+8.0})} \\
IPO & \textbf{34.50}$_{\pm 0.3}$ \small{(\textcolor{teal}{+14.9})} & 52.24$_{\pm 0.22}$ \small{(\textcolor{teal}{+21.1})} & --- & 81.35 \small{(\textcolor{teal}{+5.5})} \\
KTO & 33.81$_{\pm 0.0}$ \small{(\textcolor{teal}{+14.3})} & 51.15$_{\pm 1.77}$ \small{(\textcolor{teal}{+20.0})} & 15.31$^\ddagger$ \small{(\textcolor{teal}{+4.6})} & 80.21 \small{(\textcolor{teal}{+4.4})} \\
SimPO & 26.08$_{\pm 0.0}$ \small{(\textcolor{teal}{+6.5})} & 38.67$_{\pm 1.78}$ \small{(\textcolor{teal}{+7.5})} & 6.90$^\ddagger$ \small{(\textcolor{red}{$-$3.8})} & \textbf{85.82}$_{\pm 0.0}$ \small{(\textcolor{teal}{+10.0})} \\
\midrule
SGRPO & 32.45 \small{(\textcolor{teal}{+12.9})} & \textbf{58.00}$_{\pm 0.57}$ \small{(\textcolor{teal}{+26.8})} & --- & --- \\
GSPO & 19.86 \small{(\textcolor{teal}{+0.3})} & 39.58$^\dagger$ \small{(\textcolor{teal}{+8.4})} & --- & --- \\
CISPO & 0.15$^*$ \small{(\textcolor{red}{$-$19.4})} & 5.23$^{*\dagger}$ \small{(\textcolor{red}{$-$25.9})} & --- & --- \\
\bottomrule
\end{tabular}
\end{table}

\textbf{Cross-task generalization: MATH results.}
Table~\ref{tab:math_results} reports MATH results at 1.5B, exposing a fundamental interaction between algorithm choice and task difficulty.
On GSM8K, the spread among five methods is 19.3~pp (using seed-fix means); on MATH, this collapses to \textbf{0.54~pp}---a $36\times$ compression.
The ranking also inverts: SGRPO drops from 1st to 4th, SimPO rises from last to 2nd.
Online RL's advantage vanishes on competition-level math, confirming that algorithm choice matters most on format-sensitive tasks where structured output (not raw reasoning) drives accuracy.

\begin{table}[t]
\centering
\caption{Cross-task comparison at 1.5B: GSM8K (seed-fix mean) vs.\ MATH accuracy (\%). The 19.3~pp GSM8K spread collapses to 0.54~pp on MATH ($36\times$ compression). Rankings invert: SimPO rises from last to 2nd, SGRPO drops from 1st to 4th.}
\label{tab:math_results}
\small
\begin{tabular}{@{}lccc@{}}
\toprule
\textbf{Algorithm} & \textbf{MATH (\%)} & \textbf{GSM8K (\%)} & \textbf{$\Delta$ Rank} \\
\midrule
SFT & \textbf{27.04} & 54.36$_{\pm 0.59}$ & $\uparrow$ (2nd $\to$ 1st) \\
SimPO & 26.72 & 38.67$_{\pm 1.78}$ & $\uparrow$ (5th $\to$ 2nd) \\
KTO & 26.64 & 51.15$_{\pm 1.77}$ & $=$ (3rd $\to$ 3rd) \\
SGRPO & 26.58 & \textbf{58.00}$_{\pm 0.57}$ & $\downarrow$ (1st $\to$ 4th) \\
DPO & 26.50 & 49.08$_{\pm 0.61}$ & $\downarrow$ (4th $\to$ 5th) \\
\midrule
\textit{Spread} & \textit{0.54~pp} & \textit{19.33~pp} & \textit{$36\times$ compression} \\
\bottomrule
\end{tabular}
\end{table}

\textbf{General-domain evaluation: ARC-Challenge, HellaSwag, WinoGrande.}
To test whether these findings are specific to math or reflect a broader pattern, we evaluate all 1.5B checkpoints on three standard general-reasoning benchmarks (Table~\ref{tab:general_domain_15b}).
The result is striking: the trained-model spread collapses even further to \textbf{0.47~pp} ($41\times$ compression vs.\ GSM8K), with no method deviating more than 0.29~pp from the base model average.
SGRPO and KTO achieve the highest averages (59.48\%, 59.44\%), but neither meaningfully outperforms the untrained base (59.30\%).
SimPO, the worst performer, loses only 0.29~pp---within standard error.
This confirms that math-specific post-training neither helps nor hurts general capabilities, and that algorithm choice is entirely irrelevant on tasks outside the training distribution.

\begin{table}[t]
\centering
\caption{General-domain evaluation at 1.5B: ARC-Challenge (acc\_norm), HellaSwag (acc\_norm), WinoGrande (acc). All models trained on GSM8K. The 19.3~pp GSM8K spread collapses to 0.47~pp on general reasoning ($41\times$ compression). No method meaningfully deviates from the base model.}
\label{tab:general_domain_15b}
\small
\begin{tabular}{@{}lcccc@{}}
\toprule
\textbf{Algorithm} & \textbf{ARC-C (\%)} & \textbf{HellaSwag (\%)} & \textbf{WinoGrande (\%)} & \textbf{Avg (\%)} \\
\midrule
Base Model & 46.76 & 68.23 & 62.90 & 59.30 \\
\midrule
SFT & 46.16 \small{(\textcolor{red}{$-$0.60})} & 68.16 \small{(\textcolor{red}{$-$0.07})} & 62.90 \small{(+0.00)} & 59.08 \small{(\textcolor{red}{$-$0.22})} \\
DPO & 46.67 \small{(\textcolor{red}{$-$0.09})} & 68.38 \small{(\textcolor{teal}{+0.15})} & 62.67 \small{(\textcolor{red}{$-$0.24})} & 59.24 \small{(\textcolor{red}{$-$0.06})} \\
IPO & 46.76 \small{(+0.00)} & 68.36 \small{(\textcolor{teal}{+0.13})} & 62.67 \small{(\textcolor{red}{$-$0.24})} & 59.26 \small{(\textcolor{red}{$-$0.04})} \\
KTO & \textbf{47.01} \small{(\textcolor{teal}{+0.26})} & 68.25 \small{(\textcolor{teal}{+0.02})} & 63.06 \small{(\textcolor{teal}{+0.16})} & 59.44 \small{(\textcolor{teal}{+0.14})} \\
SimPO & 46.08 \small{(\textcolor{red}{$-$0.68})} & \textbf{68.37} \small{(\textcolor{teal}{+0.14})} & 62.59 \small{(\textcolor{red}{$-$0.32})} & 59.01 \small{(\textcolor{red}{$-$0.29})} \\
SGRPO & 46.67 \small{(\textcolor{red}{$-$0.09})} & 68.23 \small{(+0.00)} & \textbf{63.54} \small{(\textcolor{teal}{+0.63})} & \textbf{59.48} \small{(\textcolor{teal}{+0.18})} \\
\midrule
\textit{Spread (trained)} & \textit{0.93~pp} & \textit{0.22~pp} & \textit{0.95~pp} & \textit{0.47~pp} \\
\bottomrule
\end{tabular}
\end{table}

The pattern holds at 7B (Table~\ref{tab:general_domain_7b} in Appendix): despite a 9.4~pp GSM8K spread among methods (76.4--85.8\%), the general-domain spread is just \textbf{0.71~pp}.
Task-specific gains do not transfer to or harm general capabilities at either scale.

\textbf{Cross-task spread summary.}
The pattern is monotonic: as we move from the trained task (GSM8K, 19.3~pp spread) to harder reasoning (MATH, 0.54~pp, $36\times$ compression) to out-of-distribution benchmarks (general domain, 0.47~pp, $41\times$), algorithm choice becomes progressively irrelevant.
The rank correlation with GSM8K drops from 1.0 to 0.0 (MATH) to +0.6 (general domain), confirming that no method systematically transfers its advantage---or disadvantage---across tasks.

\subsection{DPO Variant Taxonomy}
\label{sec:dpo_taxonomy}

\begin{figure}[t]
\centering
\includegraphics[width=\textwidth]{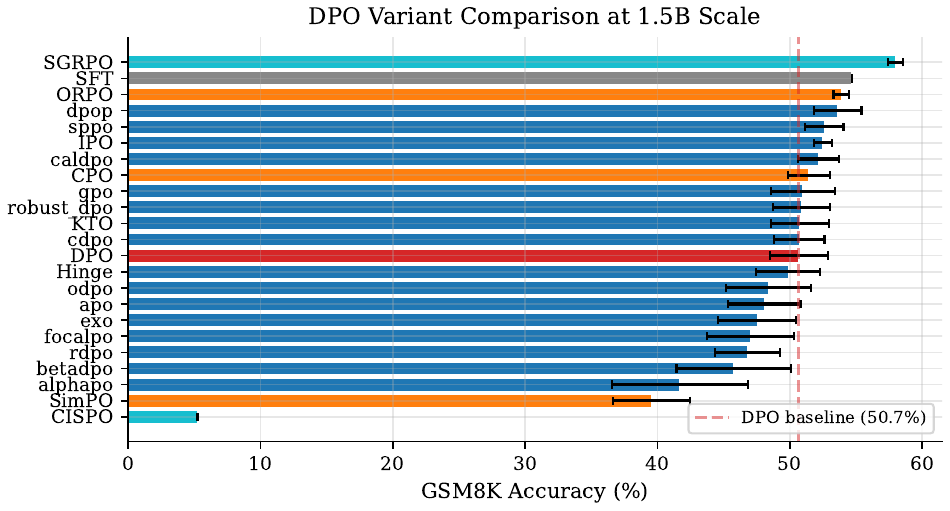}
\caption{GSM8K accuracy across 5 seeds for 20 DPO variants at 1.5B (100 runs total). Dashed line: vanilla DPO mean (49.76\%). Error bars: $\pm$1 std. $^\dagger$SimPO is the only variant significantly different from DPO ($p < 0.0026$ after Bonferroni correction)---and it is \emph{worse}.}
\label{fig:dpo_variants}
\end{figure}

We train 20 DPO variants at 1.5B with 5 seeds each (100 runs total), performing Bonferroni-corrected Welch's $t$-tests ($\alpha = 0.0026$).
The narrow spread among reference-based methods in the core comparison (0.7~pp at 0.5B, 1.2~pp at 1.5B) provides a strong prior: if most DPO loss function modifications are cosmetic, variants should cluster within a similarly narrow range.
Full results appear in Table~\ref{tab:dpo_variant_full} (Appendix~\ref{app:dpo_full}); Figure~\ref{fig:dpo_variants} visualizes the distribution.

\textbf{Zero significant winners.}
No variant achieves a statistically significant improvement over vanilla DPO (mean 49.76\% $\pm$ 2.27).
The top-ranked variant, ORPO, achieves 53.89\% $\pm$ 0.60 ($+4.12$~pp, $p = 0.013$, Cohen's $d = 2.48$)---a large effect that still fails Bonferroni correction.
The sole significant result is SimPO at 38.23\% ($-$11.54~pp, $p < 10^{-4}$, $d = -5.25$).
The 15.7~pp spread is driven entirely by methods that perform \emph{worse}; the upside is noise, the downside is real.

\textbf{Family analysis.}
No modification category yields systematic winners.
Reference-free methods span the widest range (ORPO 1st, AlphaPO 19th, SimPO last).
Grouping by divergence measure, weighting scheme, or robustness modification reveals no family-level pattern.

\textbf{Implication: loss functions are the wrong axis.}
Our 0/20 result echoes the finding of \citet{lucic2018gans} that, under controlled conditions, most GAN variants do not significantly outperform the original---despite hundreds of papers claiming otherwise.
The parallel is instructive: just as the GAN community invested enormous effort optimizing the lowest-leverage axis (generator/discriminator loss) while data and evaluation drove real gains, the preference optimization community has produced 20+ DPO variants targeting the loss function, which our hierarchy places at $\sim$1~pp impact---dwarfed by scale ($\sim$50~pp) and training paradigm ($\sim$10~pp).
Practitioners can safely use vanilla DPO; the marginal returns to loss function engineering are negligible compared to investing in better data, scale, or training paradigm choice.

\subsection{Online RL vs.\ Offline Preference}
\label{sec:online_vs_offline}

At 0.5B, SGRPO (32.45\%) slightly underperforms the best offline methods and costs $46\times$ more.
At 1.5B, the relationship reverses: SGRPO (58.0\% $\pm$ 0.57) surpasses SFT (54.4\% $\pm$ 0.59) by 3.6~pp, the largest gain of any method at this scale.
The model generates sufficient correct solutions at 1.5B to create a positive feedback loop between policy quality and training data quality.

\textbf{GRPO loss variant sensitivity.}
Token-level SGRPO outperforms sentence-level GSPO by 12.6~pp at 0.5B and $\sim$18~pp at 1.5B (where GSPO recovers to 39.58\%).
CISPO fails catastrophically at both scales (0.15\%, 5.23\%) with 73--85\% clip fractions and NaN KL divergence.
Credit assignment granularity is a high-impact choice within online RL.

\textbf{Compute efficiency.}
At 1.5B, SGRPO achieves the highest accuracy but at $\sim$6$\times$ the wall-clock cost of SFT and $10\times$ the GPU-h/pp.
At 7B, SimPO offers the best accuracy-per-GPU-hour (0.55 GPU-h/pp) alongside the highest offline accuracy.
Detailed compute tables appear in Appendix~\ref{app:compute}.

\section{Discussion}
\label{sec:discussion}

\textbf{A hierarchy of design decisions.}
Our results quantify a hierarchy of leverage: scale (${\sim}$50~pp) $\gg$ paradigm (${\sim}$10~pp) $\gg$ online/offline (${\sim}$9~pp) $\gg$ loss function (${\sim}$1~pp).
Critically, this hierarchy is \textbf{task-dependent}: the online/offline gap collapses from 8.9~pp on GSM8K to $<$0.5~pp on MATH and 0.47~pp on general-domain tasks.
The three-task trajectory (19.3~pp $\to$ 0.54~pp $\to$ 0.47~pp) confirms that algorithm choice matters only on the trained task and primarily on format-sensitive problems.

\textbf{Format compliance drives the 7B inversion.}
At 7B, SimPO's strict-match accuracy \emph{exceeds} its flexible-extract by +4.9~pp, while SFT conceals 3.9~pp of reasoning behind format-noncompliant outputs.
This pattern is absent at 1.5B (gap $<$1.4~pp).
Among preference methods at 7B, format compliance correlates inversely with reference model reliance (SimPO $>$ DPO $>$ IPO $>$ KTO; Appendix~\ref{app:format}), suggesting reference-free training encourages autonomous formatting.
This provides a mechanistic explanation for the ranking inversion: at 7B, models have sufficient capacity that format compliance---not raw reasoning---becomes the differentiating factor, and reference-free methods excel at learning formatting autonomously.

\textbf{Seed variance at 3B reveals genuine instability.}
Our seed-fix reruns at 3B expose substantial seed variance for DPO ($\sigma = 2.01$~pp across 2 seeds), indicating that small-scale post-training results are sensitive to initialization.
This contrasts with the 1.5B DPO variance ($\sigma = 0.61$~pp across 3 seeds), suggesting that post-training stability may itself exhibit scale dependence.
Combined with the hidden determinism bug in \texttt{DistributedSampler} (which made all prior SL runs byte-identical across seeds; Appendix~\ref{app:seed_invariance}), these findings underscore the need for multi-seed evaluation with verified seed propagation in all post-training studies.

\textbf{On Instruct checkpoints and DPO variant hyperparameters.}
We initialize from Instruct (not Base) by design: most practitioner post-training starts from Instruct, and using it isolates the algorithm from the base-to-instruct gap.
Preference methods are not disadvantaged: at 7B, DPO improves +8.0~pp and SimPO +10.0~pp over the Instruct base.
For the 20 DPO variants, we use each method's published default hyperparameters rather than tuning per-variant; this reflects realistic practitioner usage and avoids the combinatorial explosion of per-variant sweeps ($20 \times k$ hyperparameters).
Variant-specific tuning could recover some of the gap, but the magnitude of the null result (0/20 significant winners with $d > 2$ for the top-ranked) makes it unlikely that tuning alone would change the conclusion.

\textbf{Recommendations for practitioners.}
(1)~Validate at deployment scale; rankings at $\leq$3B do not predict 7B behavior.
(2)~At $\leq$1.5B, prefer SFT over preference methods---it is the strongest offline approach and cheapest.
(3)~Use vanilla DPO; none of 20 variants improve upon it.
(4)~At $\geq$7B with LoRA, prefer SimPO (best accuracy and compute efficiency).
(5)~For online RL, use token-level SGRPO on format-sensitive tasks where the model can self-generate correct solutions.
(6)~Always verify seed propagation in distributed training pipelines.

\section{Conclusion}

We presented the first large-scale, controlled comparison of post-training algorithms---$\sim$240 runs, 8 algorithms, 4 scales, 20 DPO variants---within the \ours{} framework.
Like \citet{lucic2018gans} for GANs, our evaluation reveals that loss function variants ($\sim$1~pp impact) target the lowest-leverage axis; scale ($\sim$50~pp) and training paradigm ($\sim$10~pp) dominate.
The central takeaway is that \emph{which algorithm is best depends on where you measure}: rankings invert across scale, across tasks, and even across evaluation metrics (strict vs.\ flexible matching).
This instability suggests the community should invest less in new loss functions and more in understanding the interaction between algorithms, scale, and task structure.

\textbf{Limitations.}
Our study uses a single model family (Qwen~2.5) up to 7B parameters.
SGRPO is absent at 3B/7B because RL requires 2~GPUs per run, limiting our factorial design.
We use synthetic preference data and greedy decoding; human-preference data and sampling-based evaluation may yield different rankings.
Code and instruction-following tasks remain untested, though we release MBPP configurations for community extension.
Despite these constraints, the consistency of our findings---replicated across four scales, three evaluation domains, 20 DPO variants with 5 seeds each, and seed-fix verification experiments at two scales---provides robust evidence for the hierarchy of leverage.
The framework's design as a \emph{living benchmark} invites the community to extend these findings to new model families, tasks, and scales.
We release all code, configs, and raw evaluation outputs for independent verification.

\bibliographystyle{plainnat}

\appendix

\section{Full DPO Variant Results}
\label{app:dpo_full}

\begin{table}[ht]
\centering
\caption{Full DPO variant results at 1.5B scale on GSM8K (5 seeds). $\Delta$ is the accuracy difference from vanilla DPO. $^\dagger$ indicates statistical significance after Bonferroni correction ($p < 0.0026$). Cohen's $d$ measures practical effect size.}
\label{tab:dpo_variant_full}
\small
\begin{tabular}{@{}llccccc@{}}
\toprule
\textbf{Variant} & \textbf{Category} & \textbf{Mean (\%)} & \textbf{Std} & \textbf{$\Delta$ DPO} & \textbf{$p$} & \textbf{$d$} \\
\midrule
ORPO & Reference-free & 53.89 & 0.60 & +4.12 & 0.0134 & 2.48 \\
DPOP & Weighting/calibration & 53.62 & 1.80 & +3.85 & 0.0189 & 1.88 \\
SPPO & Data augmentation & 52.60 & 1.45 & +2.84 & 0.0521 & 1.49 \\
IPO & Alternative divergence & 52.36 & 0.75 & +2.59 & 0.0614 & 1.53 \\
CalDPO & Weighting/calibration & 52.16 & 1.58 & +2.40 & 0.0934 & 1.22 \\
CPO & Reference-free & 51.43 & 1.57 & +1.67 & 0.2187 & 0.85 \\
GPO & Alternative divergence & 50.98 & 2.41 & +1.21 & 0.4370 & 0.52 \\
RobustDPO & Robustness & 50.89 & 2.13 & +1.12 & 0.4442 & 0.51 \\
CDPO & Robustness & 50.72 & 1.89 & +0.96 & 0.4916 & 0.46 \\
KTO & Unpaired & 50.48 & 2.11 & +0.71 & 0.6211 & 0.33 \\
Hinge & Alternative divergence & 49.87 & 2.43 & +0.11 & 0.9449 & 0.05 \\
DPO & Vanilla & 49.76 & 2.27 & -- & -- & -- \\
ODPO & Data augmentation & 48.37 & 3.20 & $-$1.39 & 0.4525 & $-$0.50 \\
APO & Data augmentation & 48.07 & 2.73 & $-$1.70 & 0.3175 & $-$0.68 \\
EXO & Data augmentation & 47.52 & 2.92 & $-$2.24 & 0.2145 & $-$0.86 \\
FocalPO & Weighting/calibration & 47.04 & 3.27 & $-$2.73 & 0.1686 & $-$0.97 \\
RDPO & Alternative divergence & 46.79 & 2.47 & $-$2.97 & 0.0835 & $-$1.25 \\
BetaDPO & Weighting/calibration & 45.75 & 4.31 & $-$4.02 & 0.1145 & $-$1.17 \\
AlphaPO & Reference-free & 41.68 & 5.13 & $-$8.08 & 0.0204 & $-$2.04 \\
SimPO$^\dagger$ & Reference-free & 38.23 & 2.12 & $-$11.54 & 0.0000 & $-$5.25 \\
\bottomrule
\end{tabular}
\end{table}

\textbf{Statistical notes.} Significance threshold: Bonferroni-corrected $\alpha = 0.05/19 = 0.0026$. Only SimPO ($p < 10^{-4}$) passes, and it is significantly \emph{worse}. ORPO ($p = 0.013$) and DPOP ($p = 0.019$) would be significant uncorrected but fail after correction. Our 5-seed design provides 80\% power to detect effects $\geq 1.0$~pp (Cohen's $d \geq 0.8$).

\textbf{Training dynamics.} Among the 20 variants, 12 produce byte-identical final loss values across all 5 seeds (deterministic, sharing the DistributedSampler behavior), while 8 exhibit genuine seed-dependent dynamics. Three variants show pathological training: SPPO diverges to loss $= 84$, CalDPO to loss $\in [10, 45]$, and BetaDPO exhibits extreme instability. Despite training pathology, SPPO still achieves 52.60\% and CalDPO reaches 52.16\%.

\section{Training Configurations}
\label{app:configs}

\begin{table}[ht]
\centering
\caption{Shared training configuration across all experiments.}
\small
\begin{tabular}{@{}ll@{}}
\toprule
\textbf{Parameter} & \textbf{Value} \\
\midrule
Optimizer & AdamW ($\beta_1\!=\!0.9$, $\beta_2\!=\!0.95$, $\epsilon\!=\!10^{-8}$) \\
Weight decay & 0.01 \\
LR schedule & Cosine with 10\% linear warmup \\
Effective batch size & 8 prompts (1 $\times$ 8 grad.\ accum.) \\
Training epochs & 3 \\
Precision & BF16 mixed precision \\
Distributed strategy & DeepSpeed ZeRO Stage 3 \\
Hardware & $8\times$ NVIDIA H100 80GB HBM3 \\
Sequence length & 512 (math), 1024 (code) \\
\bottomrule
\end{tabular}
\end{table}

\begin{table}[ht]
\centering
\caption{Algorithm-specific hyperparameters.}
\small
\begin{tabular}{@{}llll@{}}
\toprule
\textbf{Algorithm} & \textbf{Learning Rate} & \textbf{Key Hyperparameters} & \textbf{GPUs/Run} \\
\midrule
SFT & $1\!\times\!10^{-6}$ & -- & 1 \\
DPO & $1\!\times\!10^{-6}$ & $\beta = 0.1$ & 1 \\
SimPO & $1\!\times\!10^{-6}$ & $\beta = 0.1$, $\gamma = 0.5$ & 1 \\
IPO & $1\!\times\!10^{-6}$ & $\beta = 0.1$ & 1 \\
KTO & $1\!\times\!10^{-6}$ & $\beta = 0.1$ & 1 \\
SGRPO & $1\!\times\!10^{-6}$ & $\lambda_{\text{KL}} = 0$, $n = 4$ & 2 \\
GSPO & $1\!\times\!10^{-6}$ & $\lambda_{\text{KL}} = 0$, $n = 4$ & 2 \\
CISPO & $1\!\times\!10^{-6}$ & $\lambda_{\text{KL}} = 0$, $n = 4$, clip $= [0.8, 1.2]$ & 2 \\
\bottomrule
\end{tabular}
\end{table}

For models at 7B and for the 3B factorial (Table~\ref{tab:lora_2x2}), we apply LoRA~\citep{hu2022lora} with rank $r=16$, $\alpha=32$, applied to all linear layers. The core comparison at $\leq$3B uses full fine-tuning.

\section{Additional Figures}
\label{app:figures}

\begin{figure}[ht]
\centering
\includegraphics[width=\textwidth]{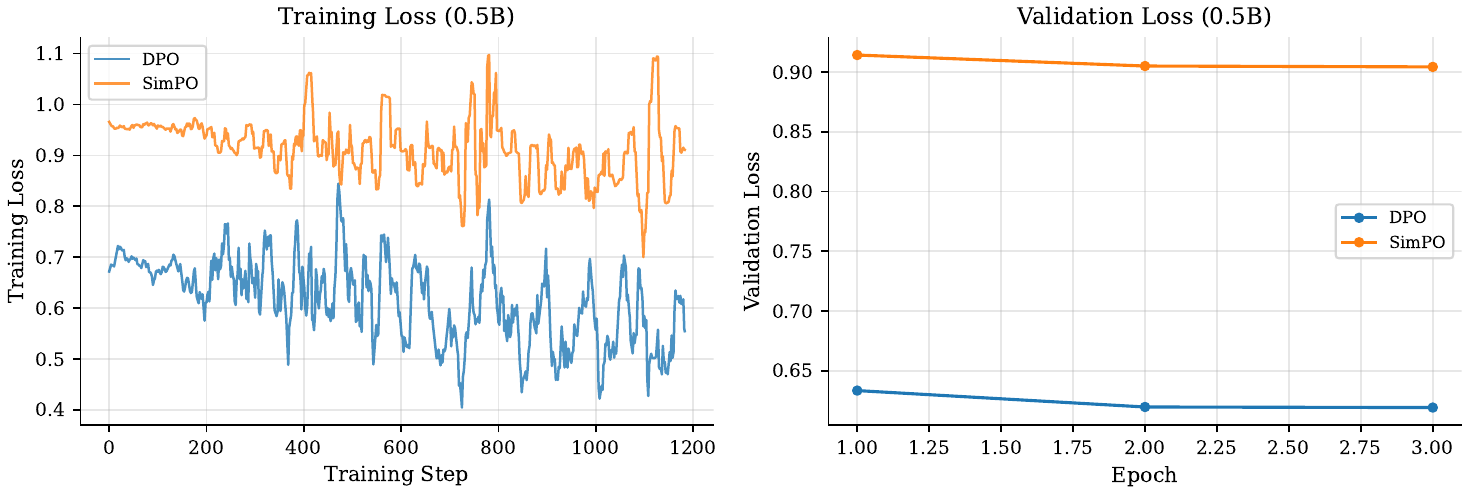}
\caption{Training dynamics comparison at 0.5B scale. \textbf{Left:} training loss over steps (smoothed). \textbf{Right:} validation loss over epochs. DPO converges to lower loss; SimPO maintains higher, noisier loss with flat validation loss.}
\label{fig:training_curves_app}
\end{figure}

\begin{figure}[ht]
\centering
\includegraphics[width=0.75\textwidth]{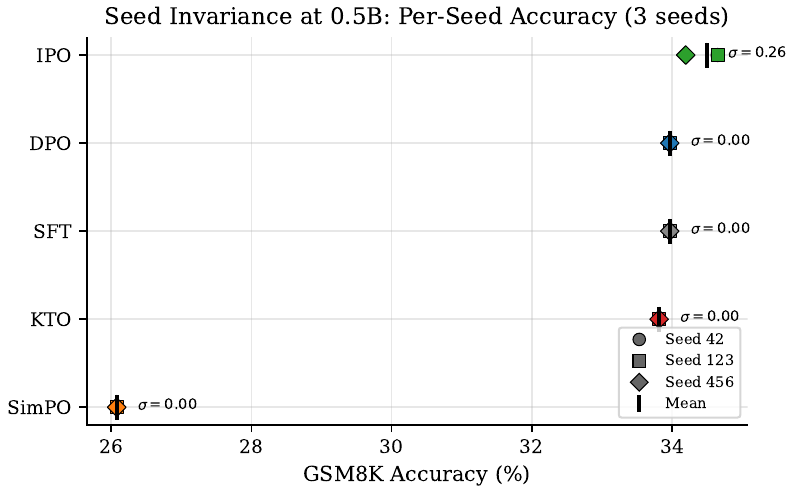}
\caption{Training determinism at 0.5B: per-seed GSM8K accuracy for each algorithm across 3 seeds. Five of six algorithms produce identical accuracy ($\sigma = 0$). Only IPO shows variance ($\sigma = 0.27$).}
\label{fig:seed_invariance_app}
\end{figure}

\section{Format Compliance Analysis}
\label{app:format}

Table~\ref{tab:format_gap} reports strict-match and flexible-extract accuracy. The \emph{format compliance gap} (strict $-$ flexible) quantifies whether a model produces well-formatted answers.

\begin{table}[ht]
\centering
\caption{Strict-match vs.\ flexible-extract accuracy on GSM8K. 3B full FT values are from original single runs (pre-seed-fix). At 7B, SimPO has the best format compliance (+4.9~pp) while SFT has the worst ($-$3.9~pp).}
\label{tab:format_gap}
\small
\begin{tabular}{@{}llccr@{}}
\toprule
\textbf{Scale} & \textbf{Algorithm} & \textbf{Strict (\%)} & \textbf{Flexible (\%)} & \textbf{Gap (pp)} \\
\midrule
\multirow{5}{*}{1.5B} & SFT & 54.66 & 55.57 & $-0.9$ \\
& DPO & 51.18 & 49.81 & $+1.4$ \\
& IPO & 52.39 & 52.31 & $+0.1$ \\
& KTO & 51.71 & 51.18 & $+0.5$ \\
& SimPO & 43.44 & 43.06 & $+0.4$ \\
\midrule
\multirow{4}{*}{3B (full FT)} & Base & 10.69 & 62.93 & $-52.2$ \\
& SFT & 18.04 & 44.28 & $-26.2$ \\
& DPO & 16.45 & 42.91 & $-26.5$ \\
& SimPO & 6.90 & 40.56 & $-33.7$ \\
\midrule
\multirow{5}{*}{7B (LoRA)} & SFT & 76.42 & 80.29 & $-3.9$ \\
& DPO & 83.85 & 84.99 & $-1.1$ \\
& IPO & 81.35 & 83.09 & $-1.7$ \\
& KTO & 80.21 & 82.49 & $-2.3$ \\
& SimPO & 85.82 & 80.97 & $+4.9$ \\
\bottomrule
\end{tabular}
\end{table}

At 3B, the base model exhibits extreme format non-compliance ($-52.2$~pp gap). Post-training partially addresses this: SFT reduces the gap to $-26.2$~pp. At 7B with LoRA, SimPO produces the most reliably formatted answers ($+4.9$~pp), while SFT shows format degradation ($-3.9$~pp). Among preference methods at 7B, the format gap correlates inversely with reference model reliance: SimPO (no ref) $>$ DPO $>$ IPO $>$ KTO, suggesting reference-free training encourages autonomous formatting.

\begin{table}[ht]
\centering
\caption{General-domain evaluation at 7B (LoRA): ARC-Challenge (acc\_norm), HellaSwag (acc\_norm), WinoGrande (acc). Despite a 9.4~pp GSM8K spread at 7B, the general-domain spread is just 0.71~pp. Rankings bear no resemblance to GSM8K rankings.}
\label{tab:general_domain_7b}
\small
\begin{tabular}{@{}lcccc@{}}
\toprule
\textbf{Algorithm} & \textbf{ARC-C (\%)} & \textbf{HellaSwag (\%)} & \textbf{WinoGrande (\%)} & \textbf{Avg (\%)} \\
\midrule
Base Model & 55.20 & 80.48 & 71.11 & 68.93 \\
\midrule
SFT & 55.29 \small{(\textcolor{teal}{+0.09})} & 80.45 \small{(\textcolor{red}{$-$0.03})} & 70.72 \small{(\textcolor{red}{$-$0.39})} & 68.82 \small{(\textcolor{red}{$-$0.11})} \\
DPO & 54.86 \small{(\textcolor{red}{$-$0.34})} & \textbf{80.58} \small{(\textcolor{teal}{+0.10})} & 70.72 \small{(\textcolor{red}{$-$0.39})} & 68.72 \small{(\textcolor{red}{$-$0.21})} \\
IPO & \textbf{55.55} \small{(\textcolor{teal}{+0.35})} & 80.37 \small{(\textcolor{red}{$-$0.11})} & \textbf{71.35} \small{(\textcolor{teal}{+0.24})} & \textbf{69.09} \small{(\textcolor{teal}{+0.16})} \\
KTO & 55.03 \small{(\textcolor{red}{$-$0.17})} & 80.39 \small{(\textcolor{red}{$-$0.09})} & 70.72 \small{(\textcolor{red}{$-$0.39})} & 68.71 \small{(\textcolor{red}{$-$0.22})} \\
SimPO & 54.44 \small{(\textcolor{red}{$-$0.77})} & 80.53 \small{(\textcolor{teal}{+0.05})} & 70.17 \small{(\textcolor{red}{$-$0.95})} & 68.38 \small{(\textcolor{red}{$-$0.56})} \\
\midrule
\textit{Spread (trained)} & \textit{1.11~pp} & \textit{0.21~pp} & \textit{1.18~pp} & \textit{0.71~pp} \\
\bottomrule
\end{tabular}
\end{table}

\section{Compute Budget}
\label{app:compute}

\begin{table}[ht]
\centering
\caption{Compute budget breakdown on H100 80GB GPUs.}
\label{tab:compute}
\small
\begin{tabular}{@{}lrrrr@{}}
\toprule
\textbf{Category} & \textbf{\# Runs} & \textbf{Time/Run} & \textbf{GPUs} & \textbf{GPU-h} \\
\midrule
Core 0.5B SL (5 algs $\times$ 3 seeds) & 15 & 4--8 min & 1 & 1.5 \\
Core 0.5B RL (3 algs $\times$ 3 seeds) & 9 & $\sim$3 h & 2 & 54 \\
Core 1.5B SL (5 algs $\times$ 3 seeds) & 15 & 14--27 min & 1 & 5.0 \\
Core 3B SL full FT (3 algs) & 3 & 12--13 min & 1 & 0.6 \\
3B LoRA factorial (3 algs) & 3 & 12--13 min & 1 & 0.6 \\
Core 7B SL (5 algs $\times$ 2 seeds) & 10 & 335--362 min & 1 & 58 \\
MATH 1.5B evals (5 methods) & 5 & $\sim$30 min & 1 & 2.5 \\
DPO variants 1.5B (20 $\times$ 5 seeds) & 100 & 14--27 min & 1 & 30 \\
Ablations 1.5B & $\sim$57 & 9--10 min & 1 & 9 \\
\midrule
\textbf{Total training} & $\sim$232 & -- & -- & $\sim$\textbf{176} \\
Evaluation (all checkpoints) & $\sim$232 & 5--30 min & 1 & $\sim$60 \\
\midrule
\textbf{Grand total} & & & & $\sim$\textbf{235 GPU-h} \\
\bottomrule
\end{tabular}
\end{table}

\begin{table}[ht]
\centering
\caption{Wall-clock training time (3 epochs) for SL methods on a single H100.}
\label{tab:compute_efficiency}
\small
\begin{tabular}{@{}lcccc@{}}
\toprule
\textbf{Algorithm} & \textbf{0.5B} & \textbf{1.5B} & \textbf{3B} & \textbf{7B (LoRA)} \\
\midrule
SFT & 4 min & 14 min & 12 min & $\sim$330 min \\
DPO & 4 min & 20 min & 13 min & $\sim$360 min \\
SimPO & 4 min & 14 min & 12 min & $\sim$330 min \\
IPO & 4 min & 26 min & -- & $\sim$335 min \\
KTO & 4 min & 23 min & -- & $\sim$460 min \\
\bottomrule
\end{tabular}
\end{table}

\begin{table}[ht]
\centering
\caption{Compute-performance Pareto frontier. GPU-h/pp = GPU-hours per percentage point gained over base. At 1.5B, SGRPO achieves highest accuracy at $10\times$ SFT's cost per point.}
\label{tab:pareto}
\small
\begin{tabular}{@{}llccccc@{}}
\toprule
\textbf{Scale} & \textbf{Method} & \textbf{GPUs} & \textbf{Time} & \textbf{Gain (pp)} & \textbf{GPU-h} & \textbf{GPU-h/pp} \\
\midrule
\multirow{3}{*}{0.5B} & \textbf{SFT/DPO} & 1 & 4 min & +14.4 & 0.07 & \textbf{0.005} \\
& IPO & 1 & 4 min & +14.9 & 0.07 & \textbf{0.004} \\
& SGRPO & 2 & 183 min & +12.9 & 6.1 & 0.47 \\
\midrule
\multirow{3}{*}{1.5B} & \textbf{SGRPO} & 2 & $\sim$84 min & +26.8 & 2.8 & 0.10 \\
& \textbf{SFT} & 1 & 14 min & +23.2 & 0.23 & \textbf{0.010} \\
& DPO & 1 & 20 min & +17.9 & 0.33 & 0.018 \\
\midrule
\multirow{3}{*}{7B} & \textbf{SimPO} & 1 & $\sim$330 min & +10.0 & 5.5 & \textbf{0.55} \\
& DPO & 1 & $\sim$360 min & +8.0 & 6.0 & 0.75 \\
& SFT & 1 & $\sim$330 min & +0.6 & 5.5 & 9.2 \\
\bottomrule
\end{tabular}
\end{table}

At $\sim$\$3/GPU-hour, the total cost is approximately \$705---modest given the breadth of comparison and substantially less than training a single 7B model from scratch. This demonstrates that comprehensive multi-seed benchmarking of post-training algorithms is accessible to groups with modest compute.

\section{Additional Scaling Results}
\label{app:scaling}

SGRPO at 1.5B achieves $58.00\% \pm 0.57$ (seeds: 57.47\%, 57.92\%, 58.61\%).
SL methods produce deterministic results across seeds (see below).
MATH results at 1.5B are in Table~\ref{tab:math_results}; MBPP configurations are released for community extension.

\section{Seed Invariance Analysis}
\label{app:seed_invariance}

\textbf{Root cause.}
Data ordering is determined by PyTorch's \texttt{DistributedSampler}, which defaults to \texttt{seed=0}. Since our framework did not pass the training seed to the sampler, all SL runs produce byte-identical parameter updates regardless of seed. Combined with deterministic BF16 gradient computation from pretrained initialization, the entire pipeline is deterministic.

\textbf{Verification.}
All SL methods produce identical evaluation accuracy across seeds at every scale (e.g., DPO at 1.5B: exactly 51.18\% for seeds 42, 123, 456). The sole exception is IPO at 0.5B, where seed 456 produces 34.19\% vs.\ 34.65\% for other seeds. DPO and SFT produce identical accuracy \emph{to each other} at 0.5B (both 33.97\%) despite different training objectives, but this parity breaks at 1.5B (SFT: 54.66\%, DPO: 51.18\%) and reverses at 7B (DPO: 83.85\%, SFT: 76.42\%).

\textbf{Post-hoc seed fix at 1.5B.}
After discovering the bug, we fixed seed propagation and retrained \emph{all five} SL algorithms at 1.5B with 3 genuine seeds each (Table~\ref{tab:seedfix}). All core findings hold: SFT (54.36\% $\pm$ 0.59) is the strongest offline method, followed by IPO (52.24\% $\pm$ 0.22), KTO (51.15\% $\pm$ 1.77), DPO (49.08\% $\pm$ 0.61), and SimPO (38.67\% $\pm$ 1.78).
Notably, IPO is the most stable method across seeds ($\sigma = 0.22$), while KTO and SimPO exhibit the highest variance ($\sigma > 1.7$).
The reference-based methods (DPO, IPO, KTO) cluster within a 3.2~pp range, consistent with the core paper's finding that loss function variants have limited impact.

\begin{table}[ht]
\centering
\caption{Seed-fix results at 1.5B on GSM8K with corrected \texttt{DistributedSampler} (3 seeds each, 15 runs total). Rankings preserved; SFT remains the strongest offline method. IPO shows the lowest variance.}
\label{tab:seedfix}
\small
\begin{tabular}{@{}lcccccc@{}}
\toprule
\textbf{Method} & \textbf{Original} & \textbf{Mean} & \textbf{$\sigma$} & \textbf{s42} & \textbf{s123} & \textbf{s456} \\
\midrule
SFT & 54.66 & 54.36 & 0.59 & 53.90 & 55.19 & 53.98 \\
IPO & 52.39 & 52.24 & 0.22 & 52.54 & 52.16 & 52.01 \\
KTO & 51.71 & 51.15 & 1.77 & 51.33 & 53.22 & 48.90 \\
DPO & 51.18 & 49.08 & 0.61 & 49.58 & 49.43 & 48.22 \\
SimPO & 43.44 & 38.67 & 1.78 & 36.85 & 38.06 & 41.09 \\
\bottomrule
\end{tabular}
\end{table}

\textbf{Post-hoc seed fix at 3B.}
We additionally ran seed-fix experiments at 3B (full FT) to verify the factorial results (Table~\ref{tab:seedfix_3b}).
DPO exhibits the largest variance ($\sigma = 2.01$~pp, seeds 16.45\% vs.\ 12.43\%), substantially higher than at 1.5B ($\sigma = 0.61$~pp), suggesting that post-training stability may itself be scale-dependent.
The 3B rankings are preserved: SFT remains the best offline method, followed by KTO, DPO, and SimPO.
Critically, SimPO remains below the base model at 3B regardless of seed, confirming that its pathological behavior at small scale is robust.

\begin{table}[ht]
\centering
\caption{Seed-fix results at 3B (full FT) on GSM8K. $^\ddagger$Single seed available. DPO shows substantially higher seed variance at 3B ($\sigma = 2.01$) than at 1.5B ($\sigma = 0.61$), suggesting scale-dependent training instability.}
\label{tab:seedfix_3b}
\small
\begin{tabular}{@{}lccccc@{}}
\toprule
\textbf{Method} & \textbf{Original} & \textbf{Mean} & \textbf{$\sigma$} & \textbf{s42} & \textbf{s123} \\
\midrule
SFT & 18.04 & 18.35$^\ddagger$ & --- & --- & 18.35 \\
KTO & --- & 15.31$^\ddagger$ & --- & --- & 15.31 \\
DPO & 16.45 & 14.44 & 2.01 & 16.45 & 12.43 \\
SimPO & 6.90 & 6.90$^\ddagger$ & --- & 6.90 & --- \\
\bottomrule
\end{tabular}
\end{table}

\end{document}